\pdfoutput=1
\documentclass[11pt]{article}

\usepackage{graphicx}
\usepackage{float}
\usepackage{authblk}
\usepackage{hyperref}
\usepackage{amsmath}

\usepackage{ACL2023}

\usepackage{times}
\usepackage{latexsym}

\usepackage[T1]{fontenc}
\usepackage[utf8]{inputenc}

\usepackage{microtype}

\usepackage{inconsolata}

\title{Benchmarking the Performance of Pre-trained LLMs across Urdu NLP Tasks}

\author{
  Munief Hassan Tahir, Sana Shams, Layba Fiaz, Farah Adeeba, Sarmad Hussain \\
  Center for Language Engineering, Al-Khawarizmi Institute of Computer Science\\ University of Engineering and Technology, Lahore, Pakistan \\
  \texttt{firstname.secondname@kics.edu.pk} 
}

\begin{document}
\maketitle

\begin{abstract}
Large Language Models (LLMs) pre-trained on multilingual data have revolutionized natural language processing research, by transitioning from languages and task specific model pipelines to a single model adapted on a variety of tasks. However majority of existing multilingual NLP benchmarks for LLMs provide evaluation data in only few languages with little linguistic diversity. In addition these benchmarks lack quality assessment against the respective state-of the art models. This study presents an in-depth examination of 7 prominent LLMs: GPT-3.5-turbo, Llama 2-7B-Chat, Llama 3.1-8B, Bloomz 3B, Bloomz 7B1, Ministral-8B and Whisper (Large, medium and small variant) across 17 tasks using 22 datasets, 13.8 hours of speech, in a zero-shot setting, and their performance against state-of-the-art (SOTA) models, has been compared and analyzed. Our experiments show that SOTA models currently outperform encoder-decoder models in majority of Urdu NLP tasks under zero-shot settings. However, comparing Llama 3.1-8B over prior version Llama 2-7B-Chat, we can deduce that with improved language coverage, LLMs can surpass these SOTA models. Our results emphasize that models with fewer parameters but richer language-specific data, like Llama 3.1-8B, often outperform larger models with lower language diversity, such as GPT-3.5, in several tasks.
\end{abstract}

\section{Introduction}

The rapid increase in the application of Artificial Intelligence (AI) across a diverse spectrum of research areas including machine translation, natural language understanding and question answering can be attributed to the remarkable performances exhibited by Foundation Models (FM) \citep{1}. Based on the framework of transformers \citep{2}, multilingual large language models (LLM) are a prominent category of foundation models that can be utilized in multiple downstream tasks. 
A number of studies have have evaluated the potential of LLMs on various Natural Language Processing (NLP) tasks. LLMRec, a LLM-based recommender system \citep{5} evaluated 3 LLMs including Llama, ChatGPT and ChatGLM on 5 recommendation tasks. \citep{6} conducted a human evaluation encompassing 10 LLMs with variations in pre-training methods, prompts, and model scales evaluated the zero-shot summarization capability. 
\citep{7} used 11 datasets covering 8 domains to evaluate the LLMs’ ability in answering common sense questions. \citep{8} conducted evaluations on 3 GPT models: ChatGPT, GPT3.5 (text-davinci-003), and text-davinci002 using 9 language pairs including low resource languages, to evaluate 18 machine translation directions. Holistic Evaluation of Language Models (HELM) project \citep{9} evaluated 30 LLMs (open, limited-access, and closed models) for English across 42 NLP tasks. \citep{10} conducted a multilingual evaluation of GPT 2.5 and Bloomz, comparing their performance with SOTA on 8 NLP tasks involving 33 languages. \citep{11} conducted a comprehensive evaluation of 214 tasks, including 48 non-English low-resource languages using 13 transformer models and 8 GPT-3 series models with varying parameters from 125 million to 175 billion. Another notable effort was conducted by \citep{12} for evaluation of 3 LLMs on 33 unique tasks for Arabic Language. 

Our study, focuses on evaluating the potential of both closed and open LLMs for supporting Urdu, a low resource language with limited data coverage in LLM's pre-training. In our experiments we utilize GPT3.5 turbo by OpenAI, Llama 2 and Llama 3.1 by Meta , Bloomz 3B and 7B1 by Big Science, Ministral 8B by Mistral AI and Whisper by OpenAI in zero-shot setting, and perform evaluation on 17 Urdu NLP tasks analyzing their performances with the existing SOTA models. To the best of our knowledge, this is the first in depth evaluation of prominent LLMs in Urdu Language context.

\begin{table*}[h]
\centering
\renewcommand{\arraystretch}{1.2} 

\resizebox{\textwidth}{!}{
\begin{tabular}{llll}
\hline
\textbf{Task} & \textbf{Dataset} & \textbf{Dataset Size} &\textbf{Testset Size}\\
\hline

Name Entity Recognition & MK-PUCIT \citep{28}& 99718 & 4165 \\
News Categorization& COUNTER \citep{21}& 1200  & 360  \\
Intent Detection& Urdu Web Queries Dataset (UWQ-22) \citep{27} & 6819  & 850  \\
Hate Speech Detection & ISE-Hate corpus \citep{17}  & 21759 & 2176 \\
Hate Speech Detection  & CLE-Hatespeech dataset \citep{18} & 5432  & 1087 \\

Propaganda Detection & ProSOUL \citep{19}& 11574 & 1737 \\

Abusive Language Detection & HASOC - Task A\citep{20} & 2400  & 240  \\
Threat Detection & HASOC - Task B\citep{20} & 9950  & 1975 \\

Cyber Bullying Identification & Cyberbullying corpus \citep{33}  & 12759 & 2480 \\

Fake News Detection & \citep{22} & 4097  & 820  \\

Hate Speech Categorization & ISE-Hate corpus\citep{17} & 8702 & 871  \\

Text Summarization  & CORPURES \citep{26} & 2649  & 311  \\

Sentiment Analysis  & \citep{29}  & 10008 & 2002 \\
Sentiment Analysis & Corpus of Aspect-based Sentiment for Urdu Political Data \citep{30} &8760 & 1450 \\

Multi-label Emotion Classification & Overview of EmoThreat (Task A) \citep{23}  & 9750  & 1950 \\

Emotion Classification             & Urdu Nastalique Emotions Dataset (UNED) \citep{25} & 4000  & 397  \\

Machine Translation(Quran) & English-Urdu Religious Parallel Corpus \citep{16}  & 6414 & 200  \\

Machine Translation(Bible)& English-Urdu Religious Parallel Corpus \citep{16}  & 7957 & 257 \\

Abstractive Summarization & CLE Meeting Corpus \citep{37}  & 240 & 10 \\

POS Tagging & Sense Tagged CLE Urdu Digest Corpus \citep{38}  & 100000 & 22522 \\

\hline

ASR (Read Speech) &  Urdu Speech Corpus \citep{36}  &-& 9.5 hours \\

ASR (Broadcast) & Urdu Broadcast (BC) Corpus\citep{35}  &-& 4.3 hours \\

\hline
\end{tabular}
}
\caption{\label{table1}
NLP Tasks and Dataset Statistics 
}
\end{table*}

\section{Approach}
For benchmarking of Urdu NLP tasks, we perform experiments using GPT 3.5, Bloomz 3B and Bloomz 7B1 , Llama 2 and Llama 3.1, Ministral 8B and Whisper in zero-shot setting and comparatively analyse the results with the respective SOTA models. Model selection was based on factors like accessibility (open/closed), infrastructure requirement, performance and language support. GPT 3.5 was selected because of its superior performance on English tasks. Among open models, popular multilingual models i.e. Llama 2 , Llama 3.1, Ministral 8B and Bloomz were evaluated for text processing tasks and Whisper models were evaluated for speech recognition task. Due to budget limitations and lack of Urdu data in the pre-training, other closed LLMs models were not investigated.

The evaluation of LLMs involved prompting and significant post-processing to extract the output in desired format. A number of prompts were curated for all NLP tasks following the recommended format and instruction pattern proposed by LAraBench \citep{12}. The prompts for each model were optimized after testing them on a few samples for each task. These prompts have been reported in Appendix.~\ref{sec:appendix} After obtaining a reasonable prompt, we used the LLM models in different settings. OpenAI's API was used for GPT 3.5.  For Bloomz, we ran the model on Google Colab utilizing 16GB GPU and for Llama 2, Llama 3.1 and Ministral 8B, we used on premises hosted versions utilizing 2X40GB A100 GPUs. Results were post-processed in all cases to align with the test set's output. The following section elaborates the LLMs (including prompting and post-processing details), NLP Tasks, Datasets,  SOTA Models and evaluation metrics, used in the study.

\subsection{Models}
\subsubsection{GPT 3.5}
GPT 3.5 Turbo has been trained on 175B parameters, encompassing both text and code data. GPT 3.5 despite being closed-source and less powerful than GPT-4 \citep{3}, is more cost-effective, as its provides free access for experimentation. Additionally, at the time of research it was the most advanced model available from OpenAI for fine-tuning.

\subsubsection{Bloomz 3B and 7.1B}
Bloomz \citep{13}, a Multitask Prompting Fine Tuned (MTF) version of the BLOOM \citep{14}, is trained on ROOTS corpus \citep{31} covering 59 languages (including 13 programming languages, and 2.59TB of Urdu language data). For evaluation, the Bloomz 3B and 7.1B models from HuggingFace were used due to their open-source availability, and optimal balance between size and computational resources. 

\subsubsection{Llama 2 and Llama 3.1}

Llama 2 \citep{32}, released by Meta, is trained on 2 trillion tokens, with 89.70\% of its content in English. 
Llama 3.1 \citep{40}, available in three variants with 8 billion, 70 billion and 405 billion parameters, is trained on over 15 trillion tokens. Both models support 8k context lengths. 
For evaluation, the Llama 2-7b and Llama 3.1-8b models were used due to their open-source availability and potential for transfer learning and generalization to languages with limited data.

\subsubsection{Ministral 8B}

The Ministral 8B \citep{42} is trained on a mixture of multilingual and code datasets, supporting a context window of up to 128k facilitated by an interleaved sliding-window attention mechanism and a vocabulary of 131k.  We benchmarked this model due to its open-source availability and its capability for low-memory inference,  and its ability to be fine-tuned and adapted to a variety of tasks.

\subsubsection{Whisper}
Whisper \citep{34}, an Automatic Speech Recognition (ASR) model developed by OpenAI, is trained on an extensive dataset comprising 680,000 hours of multilingual and multitask supervised data collected from the web. Among the diverse languages included, Whisper incorporates only 104 hours of Urdu speech corpus. For inference , we utilized the small, medium and large variants of the pre-trained Whisper model. The small variant has 12 layers, 12 attention heads, a width of 768 with 244 million parameters. The medium variant is characterized by 24 layers, 16 attention heads, a width of 1024, and consists of 769 million parameters while, the large variant features 32 layers, 20 attention heads, a width of 1280, and comprises 1550 million parameters.

\subsection{Tasks and Datasets}
This study has focused on a comprehensive evaluation of pre-trained open and closed LLMs on Urdu NLP tasks. This study utilizes 22 publicly available datasets ( see Table \ref{table1}) to evaluate 17 Urdu NLP tasks as discussed in the following sections.  

\subsubsection{Name Entity Recognition}
Name Entity Recognition (NER) is a sequence tagging task that involves identifying entities, such as names of people, organizations, locations, dates, etc. For its evaluation, we used the MK-PUCIT dataset and its SOTA model reported in \citep{28}.

% .......................................

\begin{table*}[ht]
\small 
\centering
\renewcommand{\arraystretch}{1.5} 

\resizebox{\textwidth}{!}{

\begin{tabular}{p{2.5cm} p{2.5cm} p{1.5cm} p{1.5cm} p{1.5cm} p{2cm} p{1.5cm} p{1.5cm} p{1.8cm} p{1.5cm} p{0.7cm}}
\hline

\textbf{Task} & \textbf{Dataset} & \textbf{Metric} & \textbf{GPT 3.5} & \textbf{Bloomz 3B} & \textbf{Bloomz 7B1} & \textbf{Llama 2} & \textbf{Llama 3.1}  & \textbf{Ministral 8B} & \textbf{SOTA} & \textbf{Delta} \\
\hline

Name Entity Recognition &MK-PUCIT& Macro-F1&\textbf{0.55}&0.25&0.27&0.15& 0.41 &0.25 &0.77&0.22\\

News Categorization& COUNTER & Macro-F1&\textbf{0.87}&0.58&0.48&0.13&0.64&0.67&0.70&-0.17\\

Intent Detection&Urdu Web Queries Dataset (UWQ-22)&Macro-F1&0.30&0.22&0.18&0.07&\textbf{0.42}&0.34&0.90&0.56\\

Hate Speech Detection & ISE-Hate corpus & Macro-F1 
& \textbf{0.72} & 0.52&0.53 & 0.48 &0.70&0.53& 0.83 & 0.11 \\

Hate Speech Detection & CLE-Hatespeech dataset & Macro-F1
& 0.67 &0.35 & 0.43& 0.51 &\textbf{0.72} &0.54& 0.98 & 0.26 \\

Propaganda Detection &ProSOUL&Macro-F1
&0.31&0.47&0.47&0.44&\textbf{0.66}&0.53&0.83&0.17\\

Abusive Language Detection&HAOSOC - Task A &Macro-F1&0.23&\textbf{0.51}&0.47&0.44&0.50&0.48&0.88&0.37\\

Threat Detection &HAOSOC - Task B&Macro-F1&\textbf{0.49}&0.35&0.20&0.21&0.40&0.46&0.54&0.05\\

Cyber Bullying Identification& \citep{33}&Macro-F1&0.19&0.15&0.10&0.06&\textbf{0.22}&0.08&0.84&0.41\\

Fake News Detection & \citep{22} &Macro-F1&0.55&0.52&0.51&0.47&\textbf{0.72}&0.57&0.93 &0.21\\

Hate Speech Categorization & ISE-Hate corpus  & Macro-F1
 & \textbf{0.40} & 0.28 &0.15& 0.21 &0.30&0.22& 0.83 & 0.43 \\

Extractive Summarization &CORPURES&Average Rouge-2 F1 score& 0.54&0.46&0.55&0.59&\textbf{0.62}&0.52&0.57&-0.04\\

Abstractive Summarization & CLE Meeting Corpus & Rouge-1 Score (Avg) & 0.22 & 0.02 & 0.07 & 0.006 &\textbf{0.24} &0.06& 0.31 & 0.07 \\

Sentiment Analysis& \citep{29}&Macro-F1 &\textbf{0.62}&0.35&0.33&0.3&0.44&0.36&0.88&0.26\\

Sentiment Analysis& Corpus of Aspect-based Sentiment for Urdu Political Data &Macro-F1&0.31&0.20&0.21&0.13&\textbf{0.37}&0.28 &0.70&0.37 \\

Multi-label Emotion Classification&
Overview of EmoThreat (Task A)& Macro-F1&0.20&0.17&0.26&–&\textbf{0.40}&0.29&0.68& 0.28\\

Emotion Classification & Urdu Nastalique Emotions Dataset (UNED) &Macro-F1&0.32&0.25&0.21&0.18&0.24&\textbf{0.41}&0.87&0.46\\

Machine Translation (Quran) & English-Urdu Religious Parallel Corpus & BLEU & \textbf{3.75} & 1.91 &2.36 &2.49e-78 &3.44&0.004& 13.24 & 9.49 \\

Machine Translation(Bible) & English-Urdu Religious Parallel Corpus & BLEU &5.96 & 2.28&2.47 & 0.097 &\textbf{6.43}&1.31e-78& 13.99 & 8.03 \\

POS Tagging & CLE Urdu POS Tagset & Accuracy & \textbf{0.49} & 0.11 & 0.06 & 0.09 & 0.31 &0.14& 0.96& 0.47 \\

\hline
\end{tabular}
}

\caption{Results from zero-shot experiments of GPT 3.5, Bloomz 3B, Bloomz 7B1, Llama 2, Llama 3.1 and Ministral 8B Models Compared to SOTA over NLP tasks. \textbf{Bold} text indicates the best score among models. }
\label{tab:table2}
\end{table*}

\subsubsection{News Categorization}
News categorization classify news articles into topics based on their content. For its evaluation, COUNTER dataset \citep{21} was used that consisted of articles from 5 different domains and its SOTA is reported in \citep{22}.

\subsubsection{Intent Detection}
Intent detection focuses on determining the communicative intent behind a user's input query in the form of text or speech. For our evaluation, we used the UWQ-22 dataset and SOTA model reported in \citep{27}. 

\subsubsection{Ethics and NLP: Factuality and Harmful Content Detection}

These tasks aim to evaluate the accuracy of information, identify and combat misinformation, and detect harmful content. We benchmark several tasks such as i) Hate Speech Detection using the ISE-Hate corpus by \citep{17} and CLE-Hatespeech dataset \citep{18}. ii) Propaganda Detection on the ProSOUL dataset developed by \citep{19}. iii) Abusive Language Detection in Urdu, on the dataset by \citep{20} for their Subtask A. iv) Threat Detection on the dataset of \cite{20} for Subtask B. v) Cyber Bullying Identification using Cyberbullying corpus \citep{33} vi) Fake News Detection using dataset prepared by \citep{22} vii) Hate Speech Categorization using ISE-Hate corpus by \citep{17}.

\subsubsection{Text Summarization}

Text summarization involves extracting the most important sentences from a document to create a condensed version retaining essential information. We evaluated the LLMs on:

\begin{itemize}
\item \textbf{ Extractive Summarization}

Extractive summarization condenses text by selecting and combining key sentences directly from the original content. For the evaluation of this task, we used the CORPURES dataset by \citep{26}. 

\item \textbf{ Abstractive Summarization}

Abstractive summarization generates concise summaries by understanding and paraphrasing the core meaning of a text into new, shorter sentences. For its evaluation, we have used CLE Meeting Corpus and its SOTA available in \citep{37}. 
\end{itemize}

\subsubsection{Sentiment and Emotion Analysis }
These tasks include understanding and interpreting human expressions in textual data. For Sentiment analysis, datasets from \citep{29} and CLE \citep{30} are used. For emotion analysis we used dataset from \citep{23} for their Task A: Multi-label Emotion Detection consisted of “Neutral” label and Ekman’s six basic emotions \citep{24}. The other dataset used was Urdu Nastalique Emotions Dataset (UNED) by \citep{25}. 

\subsubsection{Machine Translation}
Machine translation of Urdu is challenging due to its morphological complexity. To evaluate the translation capabilities of LLMs for English Urdu pair, we utilized the dataset by \citep{16} for Quran and Bible translations containing 200 and 257 testing samples respectively.

\subsubsection{Part of Speech (POS) Tagging}
POS tagging is a fundamental task in NLP that involves labeling each word in a sentence with its corresponding part of speech, such as noun, verb, adjective, etc. To evaluate this task we have used the CLE Urdu POS Tagset with the SOTA reported in \citep{39}. 

\subsubsection{Automatic Speech Recognition (ASR)}
ASR automatically converts spoken language into text. For its evaluation, we utilized the small, medium and large variant of the pre-trained Whisper model \citep{34}. We benchmarked this model against the SOTA models using its pre-trained weights for both broadcast and read speech recognition tasks using following two corpora:

\begin{itemize}

\item Urdu Broadcast (BC) corpus: A broadcast speech corpus \citep{35} with 4.3 hours of data from 25 speakers (14 males and 11 females). This dataset includes recordings from five different broadcast channels and YouTube, covering genres such as entertainment, health and science, current affairs, and politics.

\item Urdu Speech corpus: A read speech corpus \citep{36} consisting of 9.5 hours of Urdu speech from 62 speakers. The dataset is balanced in terms of gender and recording channels.

\end{itemize}

\begin{table*}[ht]
\small 
\centering
\renewcommand{\arraystretch}{2} 

\resizebox{\textwidth}{!}{
\begin{tabular}{p{1.5cm} p{2.5cm} p{1.5cm} p{1.5cm} p{1.5cm} p{1cm} p{1cm} p{1cm}}
\hline

\textbf{Task} & \textbf{Domain} & \textbf{Metric} & \textbf{Whisper (Large)} & \textbf{Whisper (Medium)} & \textbf{Whisper (Small)} & \textbf{SOTA} & \textbf{Delta}   \\
\hline

ASR & Read Speech & WER & \textbf{23.51} & 27.88 & 36.90 &  16.94  & -6.57  \\

ASR & Broadcast & WER & \textbf{27.97} & 35.57 &42.57& 18.59  &   -9.38  \\
\hline
\end{tabular}
}
\caption{Performance Metrix of ASR for Whisper Large, Medium and Small Models Compared to SOTA.}
\label{tab:table3}
\end{table*}

\subsection{Zero-Shot Setup}
For all LLMs; GPT 3.5, Bloomz 3b and 7b, Llama 2 and Llama 3.1 and Ministral 8B we use zero-shot prompting giving natural language instructions describing the task and specify the expected output. Prompts allow LLMs to learn context and narrows the inference space to produces accurate output as further elaborated in the section \ref{subsec: prompt engineering}.

\subsection{Inference Settings}
The inference experiments for Llama 2, Llama 3.1 and Ministral 8B were conducted using two parallel NVIDIA A100-PCIE-40GB GPUs, providing a combined computational capacity of 80GB. During the inference, nearly 90 percent of the total GPU capacity was utilized. For experiments of GPT-3.5, API from OpenAI was utilized. Inference experiments with GPT-3.5 were conducted using Google Colab. Inference experiments with Bloomz's 3B and 7.1B models, available on huggingface,  were also conducted using Google Colab. For Speech processing experiments using Whisper, two NVIDIA RTX3060-12GB GPUs were employed, providing a combined computational capacity of 24GB.

\subsection{Prompt Engineering and Post Processing}
\label{subsec: prompt engineering}

In our experimentation with different LLMs, we tweaked the prompts based on the models input. Prompts for tasks such as News categorization \ref{subsec:news categorization}
and Hate speech Categorization \ref{subsec:hatespeech categorization} were challenging because they required outputs from pre-defined ground-truth categories. Prompts for Machine Translation task \ref{subsec:MT} had to be engineered so that the model's output only includes the translated text. Thus optimal prompts were curated by testing against each model on few samples, while ensuring no bias in decision-making.

Despite careful prompting, model responses required post-processing to align with desired outcomes e.g. capitalization ("fake" vs. "Fake"), standardizing output formats ("1. Propaganda" to "1"), and omitting "explanations" and "note" produced with the models' responses, specifically in Hate speech detection \ref{subsec:hatespeech detection} task.
Some model outputs didn't match desired outcomes, e.g. News categorization included 5 domains i.e. sports, showbiz, foreign , national , business however the models output out of context domains such as "politics" and "entertainment". Among all the models, Llama 2 required the most output post-processing.

For a thorough description of the prompts crafted for each LLM, please refer to Appendix \ref{sec:appendix}.

\subsection{SOTA Models}
In this study, we benchmark the capabilities of LLMs in a zero-shot scenario by comparing them with SOTA models as reported in respective studies. These SOTA models employed diverse architectures including Capsule NN, Support Vector Machine (SVM), Random Forest (RF), Decision Tree (J48), Sequential Minimal Optimization (SMO), Convolutional Neural Networks (1D-CNN), LSTM with CNN features , Naive Bayes classifier and various multilingual transformer models such as m-BERT and frameworks like XGboost and LGBM.

\subsection{Evaluation Metrics}
The evaluation metrics used for the experiments have been kept identical to the one used in the respective state of the art references. They are Macro-F1, Rouge 2 F1 score, BLEU
\footnote{\url{https://www.nltk.org/api/nltk.translate.bleu
}}, accuracy and Word Error Rate (WER).   We have also computed the delta to highlight the differential between best performing LLM's output with the SOTA model.

\section{Results and Discussion}
\begin{figure*}[!ht]
    \vspace{-2em}  % Adjust this value as needed
    \centering
    \includegraphics[width=\textwidth]{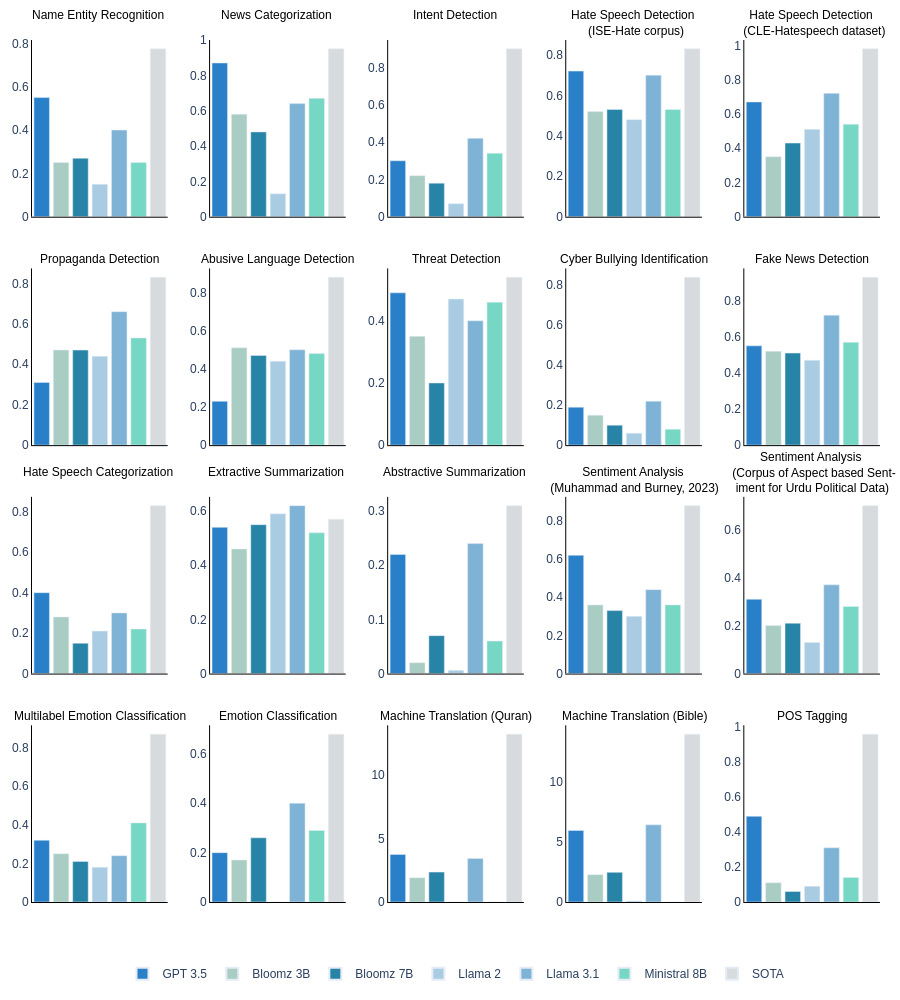} 
    \caption{The performance of different models in zero-shot scenario as compared to SOTA. Missing bars in some tasks mean that the specific model cannot perform the specified task.}
    \label{fig:result}
    \vspace{0.2em}  % Adjust this value as needed
\end{figure*}

The results on text processing tasks of our experimentation have been summarized in Figure \ref{fig:result}. The Figure presents a grid of bar graphs for each NLP task, with the y-axis showing evaluation metrics specific to each task. For classification and detection tasks, the y-axis represents the macro F1 score. For summarization tasks, it shows the average ROUGE-2 score, while for machine translation tasks, it displays the BLEU score. Each model is represented by a distinct color bar , as indicated in the Figure's legend, which is kept consistent across all tasks, with the bar of SOTA providing a reference point for comparison. Missing bars in certain tasks indicate that the model outputs were effectively zero (e.g., in Table \ref{tab:table2} value is 2.49e-78 for Llama 2 on the Machine Translation (Quran) task, and 1.31e-78 for Mistral 8B on the Machine Translation (Bible) task), reflecting negligible performance.

Our results show that LLMs differ in their applicability to different data regimes and tasks. LLM models were able to surpassed the SOTA model for news categorization with GPT 3.5 and Llama 3.1 for Extractive Summarization. In all other experiments, LLMs remained lower than the SOTA models (reference Table \ref{tab:table2}). Across all experiments, Llama 3.1 outperformed in 10 of the 17 tasks, while GPT-3.5 excelled in 8 tasks. In comparison, Bloomz and Ministral 8B each led in only one task. The minimum delta obtained was 0.05 between GPT 3.5 and SOTA model for threat detection task.  In comparison with other the open LLMs, Llama 3.1 performed better in majority of the NLP tasks which is due to its extensive multilingual data, architecture and advanced training techniques, enabling it to effectively generalize across languages and tasks.

In choice and evaluation of LLMs, Bloomz 3B and Bloomz 7B1 were initially chosen for experimentation due to their early introduction and multilingual capabilities. However, they have not kept pace with advancements seen in other models like Llama. Analysis reveals that there is no significant performance efficiency gained from transitioning from Bloomz-3B to Bloomz-7B1 as evident from Table \ref{tab:table2}. In contrast, the performance of Llama models has notably improved, particularly from Llama 2 to Llama 3.1, indicating a more effective evolution in their design and capabilities.

Based on our evaluations, the top two performing models for NLP tasks are GPT-3.5 and Llama 3.1 with comparable performances as evident from Figure \ref{fig:result}. Llama 3.1 outperformed GPT 3.5 in 11 NLP tasks and was even better than SOTA in Extractive Summarization. On the other hand, GPT 3.5 was better than Llama 3.1 in 8 tasks and surpassed SOTA in News Categorization task. Llama 3.1's superior performance is due to the increased coverage of Non-English data in the model as well as increased amount of pretraining data i.e. 15 trillion tokens. 

The performance of Ministral 8B is comparable to GPT 3.5 and Llama 3.1. It outperformed GPT 3.5 in 6 NLP Tasks i.e. Intent Detection, Propaganda Detection, Abusive Language Detection, Fake News Detection, Multi Label Emotion Classification and Emotion Classification. And outperformed Llama 3.1 in News Categorization and it was best among all models in Emotion Classification. However its performance was quite inadequate on generation tasks like Machine Translation (reference Figure \ref{fig:result}). Overall its performance is good in detection tasks and is attributed to its interleaved sliding-window attention mechanism, which enables it to efficiently handle extended contexts with reduced memory usage. Detection tasks such as Fake News Detection and Propaganda Detection often require recognizing patterns across longer texts. This enhanced ability to retain and utilize extended contextual information allowed Ministral 8B to excel in detection tasks.

The results of Speech Processing tasks are summarized in Table \ref{tab:table3}. The analysis of the results indicates that the SOTA models, which were trained on a larger corpus of Urdu data, outperformed all variants of the Whisper model across the evaluated datasets. These findings suggest that while the SOTA models trained on more extensive Urdu datasets exhibited superior performance, the larger variant of the Whisper also demonstrated improved performance compared to its medium and small counterpart, underscoring the importance of model size and complexity in ASR tasks. The negative delta values indicate that the model's performance falls below the SOTA benchmarks, highlighting a gap to be addressed.

Error analysis of the LLMs' output against the ground truth revealed two main factors that account for the decline in overall F1 scores of LLMs. The factors include i) discrepancies in the output format, where the output contained extra or omitted tokens, and ii) the generation of out-of-scope labels. These observations imply that the seamless deployment of LLMs may be challenging, requiring substantial efforts either in formulating precise prompts for accurate outputs or engaging in post-processing to align the outputs with reference labels.

Thus, performance of LLMs significantly depends on well-curated prompts and intelligent post-processing of the outputs. While Llama 2 and Bloomz show a notable performance deficit compared to the SOTA, the newer Llama version i.e. Llama 3.1 and GPT 3.5 succeeds in mitigating this gap to a considerable extent. 

\section{Conclusion and Future Work}
\label{sec:conclusion}
In this study, we benchmark the potential of both open and closed LLMs on 17 Urdu NLP tasks employing a substantial number of publicly accessible datasets. Through our experiments we provide a comparative performance analysis for each task and dataset against the SOTA. These findings will assist the Urdu NLP community in selecting suitable models for usage and fine-tuning within specific contexts. As future work, we aim to develop a public leader board for Urdu benchmarking and explore integration of additional models, tasks, and datasets. Also, after evaluating multiple models, we are focusing on pretraining Llama 3.1 to enhance Urdu language support and expand token coverage for greater adaptability across diverse languages and domains. This will also includes domain-specific fine-tuning to further boost performance, leveraging Llama 3.1’s compact size, active community, and robust results.

\section*{Limitations}
Our study is confined to seven LLMs and does not include the heavier versions of models such as Bloomz-170B or Llama 3.1 405B due to hardware and computational resource limitations which may impact the comprehensiveness of the analysis. This limitation may affect the generalization of the findings to models with higher parameters, potentially missing insights into the performance of more robust versions of these language models. Our study also primarily concentrates on evaluating the models in a zero-shot setting. While this setting provides valuable insights into the models' out-of-the-box performance, it may not capture the full potential of fine-tuned models for specific tasks. Our study also does not extensively delve into the quality and representativeness of the training data for Urdu language used in these models. 

\section*{Acknowledgments}

We would like to express our sincere gratitude to Dr. Miriam Butt for allowing us to utilize the infrastructure at the University of Konstanz for our benchmarking experiments. Additionally, we extend our acknowledgment to Miss Ayesha Khalid for supplying the ASR dataset and corresponding SOTA techniques.

\bibliography{anthology,custom}
\bibliographystyle{acl_natbib}

\appendix

\section{Appendix}
\label{sec:appendix}

\subsection{Prompts - Name Entity Recognition} 
\subsubsection{Bloomz} 
Perform Name Entity Recognition for the words using the following technique:
- Mark names, nicknames, cast, family, and relational names as Person.
- Mark names of companies, media groups, teams, and political parties as Organization.
- Mark all man-made structures and politically defined locations, such as names of countries, cities, and places like railway stations, as Location.
- Mark all remaining words, such as prepositions, adjectives, adverbs, and names of books and movies, as Other.
No explanation is required. Just output the Entity name.Word: {} Entity:

\subsubsection{GPT 3.5 } 
Perform Name Entity Recognition corresponding to each word using the following annotation technique:
Person : name,nickname,cast,family,relational names and titles. God's name should NOT be marked as Person.
Organization : name of company, media group, team,political party. Name of product or brand should NOT be marked as Organization.
Location : all man-made structures and politically defined locations such as names of countries,city and places like railway station etc. A generic reference to location should NOT be marked as Location.
Other : all remaining words, such as prepositions, adjectives, adverbs, names of books and movies etc.
No explanation is required. Just output the tag name.
word = {}

\subsubsection{Llama 2 }
You are Performing Name Entity Recognition for the urdu words.<</SYS>>
Human:
Word: {}
Please select one of the following entity:
Person
Organization
Location
Other
No explanation or further assistance is required. Only entity name is required
Assistant:
The entity is

\subsubsection{Llama 3.1 }
You are a name entity recognition model. Your task is to mark the entity as Person, Organization, Location, or Other in Urdu text samples. Ensure that your outputs are Person, Organization, Location, or Other. No explanation is required.

\subsubsection{Ministral 8B}
Perform Name Entity Recognition for the words using the following technique:
- Mark names, nicknames, cast, family, and relational names as Person.
- Mark names of companies, media groups, teams, and political parties as Organization.
- Mark all man-made structures and politically defined locations, such as names of countries, cities, and places like railway stations, as Location.
- Mark all remaining words, such as prepositions, adjectives, adverbs, and names of books and movies, as Other.
No explanation is required. Just output the Entity name.Word: {} Entity:

\subsection{Prompts - News Categorization}
\label{subsec:news categorization}

\subsubsection{Bloomz}
News: {}
Classify the given news into one of the following category
0. sports
1. national
2. foreign
3. showbiz
4. business
Choose the best suited label from above. Your output should be 0-4 only. No explanation. Only 0-4. No other label or additional text.
Label (0,1,2,3,4):

\subsubsection{GPT 3.5 }
News: {}
Classify the given news into one of the following category
0. sports
1. national
2. foreign
3. showbiz
4. business
Choose the best suited label from above. Your output should be the name of the category only. No explanation.No other label or additional text. Category:

\subsubsection{Llama 2 }
Provide the label of the above news from the following:
0. sports
1. national
2. foreign
3. showbiz
4. business
No explanation. Please answer in numbers
News : {}
Answer:

\subsubsection{Llama 3.1 }
News: {}
Classify the given news into one of the following category
0. sports
1. national
2. foreign
3. showbiz
4. business
Choose the best suited label from above. Your output should be 0-4 only. No explanation. Only 0-4. No other label or additional text.
Label (0,1,2,3,4):

\subsubsection{Ministral 8B}
News: {}
Classify the given news into one of the following category
0. sports
1. national
2. foreign
3. showbiz
4. business
Choose the best suited label from above. Your output should be 0-4 only. No explanation. Only 0-4. No other label or additional text.
Label (0,1,2,3,4):

\subsection{Prompts - Intent Detection} 
\subsubsection{Bloomz}
You are an intent classification model. Your task is to identify the intent in the following urdu sentence. {}
Intents are:
0. Informational
1. Navigational
2. Transitional
Output (0,1,2):

\subsubsection{GPT 3.5 }
"system": "You are an intent detection classification model.
You are an intent classification model. Your task is to identify the intent in the following urdu sentence. {}
Intents are:
0. Informational
1. Navigational
2. Transitional
Output (0,1,2):

\subsubsection{Llama 2}
You are an intent classification model. Your task is to identify the intent in the following urdu sentence. {}
Intents are:
0. Informational
1. Navigational
2. Transitional
Dont write any explanation or reason for answer.
Output (0,1,2):

\subsubsection{Llama 3.1}
You are an intent classification model. Your task is to mark the intent as 0 or 1 or 2 in Urdu text samples. Ensure that the model outputs '0' for Informational intent , '1' for Navigational intent and '2' for  Transitional intent. Ensure that your outputs 0 or 1 or 2 only. No explanation is required.

\subsubsection{Ministral 8B}
You are an intent classification model. Your task is to identify the intent in the following urdu sentence. {}
Intents are:
Informational
Navigational
Transitional
Only output the name of the intent. No explanation is required.

\subsection{Prompts - Hate Speech Detection ISE-Hate corpus}
\label{subsec:hatespeech detection}
\subsubsection{Bloomz}
Classify the hate sentence into the category it falls:
Ethnic
Interfaith
Sectarian
Other
Output "0" for Other, "1" for "Sectarian", "2" for "Interfaith" and "3" for "Ethnic"
Sentence: {}
Class:

\subsubsection{GPT 3.5 }
"system": "You are an expert in detecting hate speech in the urdu samples "
Classify the hate sentence into the category it falls:
Ethnic
Interfaith
Sectarian
Other
Output "0" for Other, "1" for "Sectarian", "2" for "Interfaith" and "3" for "Ethnic". No explanation is required
Sentence: {} Output (0,1,2,3):

\subsubsection{Llama 2}
You are a hate speech classification model.
Labels:
1: Sectarian hate
2: Interfaith hate
3: Ethnic hate
0: None of the above
Instructions: To distinguish between hate speech and non-hate speech in text samples. Ensure that the model outputs "1" for hate related to "Sectarian", "2" for hate related to "Interfaith" and "3" for hate related to "Ethnic" and "0" if you think it does not fall in these three categories.
Your output should be only 0, 1, 2 or 3. No explanation is required.
Sentence: {}
Label(0,1,2,3):

\subsubsection{Llama 3.1}
You are a hate speech classification model. Your task is to mark 0 or 1 or 2 or 3 in Urdu text samples. Ensure that the model outputs '1' for Sectarian hate text , '2' for Interfaith hate text , '3' for Ethnic hate text and '0' for any other hate text. Ensure that your outputs 0 or 1 or 2 or 3 only. No explanation is required.

\subsubsection{Ministral 8B}
You are a hate speech classification model.
Labels:
1: Sectarian hate
2: Interfaith hate
3: Ethnic hate
0: None of the above
Instructions: To distinguish between hate speech and non-hate speech in text samples. Ensure that the model output "1" for hate related to "Sectarian", "2" for hate related to "Interfaith" and "3" for hate related to "Ethnic" and "0" if you think it does not fall in these three categories.
Your output should be only 0, 1, 2 or 3. No explanation is required. 
Sentence: {}
Label(0,1,2,3):

\subsection{Prompts - Hate Speech Detection CLE corpus }
\label{subsec:hatespeech detection}

\subsubsection{Bloomz}
Classify the sentence as hate speech or non-hate speech. Output "1" for non-hate speech and "-1" for hate speech. No explanation is required. Sentence: {} Output(1 or -1):

\subsubsection{GPT 3.5 }

"system": "You are an expert in detecting hate speech in the urdu samples"
Classify the sentence as hate speech or non-hate speech. Output "1" for non-hate speech and "-1" for hate speech. No explanation is required. Sentence: {}

\subsubsection{Llama 2}
You are a hate speech classification model.
Labels:
1: Non-hate speech
-1: Hate speech
Instructions: To distinguish between hate speech and non-hate speech in text samples. Ensure that the model outputs 1 for non-hate speech and -1 for hate speech.
Your output should be only 1 or -1. No explanation is required
Sentence: {}
Label(1 or -1):

\subsubsection{Llama 3.1}
You are a hate speech classification model. Your task is to mark 0 or 1 in Urdu text samples. Ensure that the model outputs '0' for Non-hate text and '1' for Hate text. Ensure that your outputs 0 or 1 only. No explanation is required.

\subsubsection{Ministral 8B}
You are a hate speech classification model.
Labels:
0: Non-hate speech
1: Hate speech
Instructions: To distinguish between hate speech and non-hate speech in text samples. Ensure that the model outputs 0 for non-hate speech and 1 for hate speech.
Your output should be only 0 or 1
Sentence: {}
Label:

\subsection{Prompts - Propaganda Detection}
\subsubsection{Bloomz}
Classify the article as Propaganda or Non-Propaganda. Output '1' for Propaganda and '0' for Non-Propaganda. Don't concatenate input with output. No explanation is required. The article is: {}. Class(0 or 1):

\subsubsection{GPT 3.5 }
Classify the article as Propaganda or Non-Propaganda. Output '1' for Propaganda and '0' for Non-Propaganda. No explanation is required. The article is: {}. Class(0 or 1):

\subsubsection{Llama 2}
Classify the article as Propaganda or Non-Propaganda. Output '1' for Propaganda and '0' for Non-Propaganda. Don't concatenate input with output. No explanation is required. The article is: {}. Class(0 or 1):

\subsubsection{Llama 3.1}
Classify the article as Propaganda or Non-Propaganda. Output '1' for Propaganda and '0' for Non-Propaganda. Don't concatenate input with output. No explanation is required. The article is: {}. Class(0 or 1):

\subsubsection{Ministral 8B}
Classify the article as Propaganda or Non-Propaganda. Output '1' for Propaganda and '0' for Non-Propaganda. No explanation is required. The article is: {}. Class (0 or 1):

\subsection{Prompts - Abusive Language Detection}
\subsubsection{Bloomz} 
You are an abusive language detection model.
Labels:
0: non-abusive language
1: abusive language
Instructions: To distinguish between abusive and non-abusive language in text samples. Ensure that the model outputs 0 for non-abusive language and 1 for abusive language.
Your output should be only 0 or 1
Sentence: {}
Label:

\subsubsection{GPT 3.5 }
"system": "You are an expert in detecting abusive language in the urdu samples
Classify the sentence as abusive language or non-abusive language. Output "1" for non-abusive language and "0" for abusive language. No explanation is required. Sentence: {}

\subsubsection{Llama 2 }
You are a abusive language detection model.
Labels:
0: non-abusive language
1: abusive language
Instructions: To distinguish between abusive and non-abusive language in text samples. Ensure that the model outputs 0 for non-abusive language and 1 for abusive language.
Your output should be only 0 or 1. No explanation
Sentence: {}
Label(0 or 1):

\subsubsection{Llama 3.1 }
You are a abusive language detection model.

Labels:

0: non-abusive language
1: abusive language
Instructions: To distinguish between abusive and non-abusive language in text samples. Ensure that the model outputs 0 for non-abusive language and 1 for abusive language.
Your output should be only 0 or 1. No explanation
Sentence: {}
Label(0 or 1):

\subsubsection{Ministral 8B}

You are a abusive language detection model.
Labels:
0: non-abusive language
1: abusive language
Instructions: To distinguish between abusive and non-abusive language in text samples. Ensure that the model outputs 0 for non-abusive language and 1 for abusive language.
Your output should be only 0 or 1. No explanation
Sentence: {}
Label(0 or 1):

\subsection{Prompts - Threat Detection} 

\subsubsection{Bloomz}
Classify the sentence as threatening or non threatening. Output class "1" for threatening and "0" for non threatening.  Sentence: {} Class(1 or 0):

\subsubsection{GPT 3.5 }
system: You are an expert in detecting threat in the urdu samples
Classify the sentence as threatening or non threatening. Output "1" for threatening and "0" for non threatening.  Sentence: {}:

\subsubsection{Llama 2}
Classify the sentence as threatening or non threatening. Output class "1" for threatening and "0" for non threatening. No explanation required. Sentence: {} Output(1 or 0):

\subsubsection{Llama 3.1}
You are a classification model. Your job is to classify the sentences as threatening or non-threatening. Output "1" if the sentence is threatening and "0" if it is non-threatening. No explanation is required

\subsubsection{Ministral 8B}
Classify the sentence as threatening or non threatening. Output "1" for threatening and "0" for non threatening. No explanation is required. Sentence: {} 
\subsection{Prompts - Cyber Bullying Identification}
\subsubsection{Bloomz}
Your task is to classify the nature of cyberbullying with one of the labels:
INSULT
OFFENSIVE
NAMECALLING
PROFANE
THREAT
CURSE
NONE
Output only label name. no explanation is required. Sentence {} . Output label:

\subsubsection{GPT 3.5 }
Your task is to classify the nature of cyberbullying with one of the labels:
INSULT
OFFENSIVE
NAMECALLING
PROFANE
THREAT
CURSE
NONE
Output only label name. no explanation is required. Sentence {} . Output label:

\subsubsection{Llama 2}
You are a helpful assistant in classification of cyberbullying. You should always provide answer from given labels without explanation.
<</SYS>>
Human:
Sentence {}.  classify the nature of cyber bullying present in sentence with one of the following label:
INSULT
OFFENSIVE
NAMECALLING
PROFANE
THREAT
CURSE
NONE
Assitant:

\subsubsection{Llama 3.1}
You are a cyberbullying classification model. Your task is to mark the input as INSULT or OFFENSIVE or NAMECALLING or PROFANE or THREAT or CURSE or NONE in Urdu text samples. Ensure that your output is INSULT or OFFENSIVE or NAMECALLING or PROFANE or THREAT or CURSE or NONE. No explanation is required.

\subsubsection{Ministral 8B}

Your task is to classify the nature of cyberbullying with one of the labels:
INSULT
OFFENSIVE
NAMECALLING
PROFANE
THREAT
CURSE
NONE
Output only label name. no explanation is required. Sentence {} . Output label:

\subsection{Prompts - Fake News Detection}
\subsubsection{Bloomz}
You are a fake news detection model.
Labels:
fake
real
Instructions: To distinguish between fake news and real news in text samples. Ensure that the model outputs 'fake' for fake news and 'real' for real news.
No explanation is required
Sentence: {}
Label(fake or real):

\subsubsection{GPT 3.5 }

"system": "You are an expert in detecting fake news in the urdu samples”
You are a fake news detection model.
Labels:
fake
real
Instructions: To distinguish between fake news and real news in text samples. Ensure that the model outputs 'fake' for fake news and 'real' for real news.
No explanation is required
Sentence: {}
Label(fake or real):

\subsubsection{Llama 2}
You are a fake news detection model.
Labels:
fake
real
Instructions: To distinguish between fake news and real news in text samples. Ensure that the model outputs 'fake' for fake news and 'real' for real news.
No explanation is required
Sentence: {}
Label(fake or real):

\subsubsection{Llama 3.1}
You are a fake news detection model.Output "fake" for fake news and "real" for real news. No explanation is required.

\subsubsection{Ministral 8B}

You are a fake news detection model.
Labels:
fake
real
Instructions: To distinguish between fake news and real news in text samples. Ensure that the model outputs 'fake' for fake news and 'real' for real news.
No explanation is required
Sentence: {}
Label(fake or real):

\subsection{Prompts - Hate Speech Categorization}
\label{subsec:hatespeech categorization}

\subsubsection{Bloomz}
You are a hate speech classification model.
Labels:
0: Non-hate speech
1: Hate speech
Instructions: To distinguish between hate speech and non-hate speech in text samples. Ensure that the model outputs 0 for non-hate speech and 1 for hate speech.
Your output should be only 0 or 1
Sentence: {}
Label:

\subsubsection{GPT 3.5 }
You are a hate speech classification model.
Labels:
0: Non-hate speech
1: Hate speech
Instructions: To distinguish between hate speech and non-hate speech in text samples. Ensure that the model outputs 0 for non-hate speech and 1 for hate speech. Your output should be only 0 or 1
Sentence: {}

\subsubsection{Llama 2}
You are a hate speech classification model.
Labels:
0: Non-hate speech
1: Hate speech
Instructions: To distinguish between hate speech and non-hate speech in text samples. Ensure that the model outputs 0 for non-hate speech and 1 for hate speech.
Your output should be only 0 or 1. No explanation is required
Sentence: {}
Label(0 or 1):

\subsubsection{Llama 3.1}

You are a hate speech classification model. Your task is to mark 0 or 1 in Urdu text samples. Ensure that the model outputs '0' for Non-hate text and '1' for Hate text. Ensure that your outputs 0 or 1 only. No explanation is required.

\subsubsection{Ministral 8B}
You are a hate speech classification model. Your task is to mark 1 or -1 in Urdu text samples. Ensure that the model outputs '1' for Non-hate text and '-1' for hate text. Ensure that your outputs 1 or -1 only. No explanation is required. {}

\subsection{Prompts - Extractive Summarization}
\subsubsection{Bloomz}
You are an extractive summarization model. Label the sentence that you considered is important for Summarization as "1". If you think sentence should not be kept for extractive summary, label it as "0". Sentence {} Label:

\subsubsection{GPT 3.5 }
"system": "You are an extractive summarization model for Urdu language"
You are an extractive summarization model. Label the sentence that you considered is important for Summarization as "1". If you think sentence should not be kept for extractive summary, label it as "0". Sentence {} Label:

\subsubsection{Llama 2}
Passage: {} For extractive summarization, should this passage be kept or discarded? Act as a summarization model. Provide answer only (0 or 1) without explanation. Answer:

\subsubsection{Llama 3.1}
You are an extractive summarization model. You have to decide whether the given passage should be kept or discarded for summary? Output "1" if the passage should be kept and "0" for discarding (0 or 1) without explanation

\subsubsection{Ministral 8B}
You are an extractive summarization model. You have to decide whether the given passage should be kept or discarded for summary? Output "1" if the passage should be kept and "0" for discarding (0 or 1) without explanation

\subsection{Prompts - Abstractive Summarization}

\subsubsection{Bloomz}
Write summary of the given Urdu meeting. Meeting: {}. Summary:

\subsubsection{GPT 3.5 }
You are a summarization model. Generate summary for the given meeting minutes in Urdu.
Prompt: Meeting minutes: {} Summary:

\subsubsection{Llama 2}
You are a summarization model. Your job is to summarize the urdu meeting minutes.
Meeting minutes: {} Summary:

\subsubsection{Llama 3.1}
You are a summarization model. Generate the summary for the meeting minutes in Urdu.

\subsubsection{Ministral 8B}
You are a summarization model. Generate the summary for the meeting minutes in Urdu.

\subsection{Prompts - Sentiment Analysis}
\subsubsection{Bloomz}
Do the sentimental analysis. Output should be "pos" for positive sentence , "neu" for neutral sentence and "neg" for negative sentence. No explanation is required.
Sentence: {}
Label:

\subsubsection{GPT 3.5 }
Do the sentiment analysis. Output should be "pos" for positive sentences , "neu" for neutral sentences and "neg" for negative sentences. No explanation is required. Sentence: {}

\subsubsection{Llama 2}

You are a helpful assistant in sentiment analysis. You should always provide answer from given labels without explanation.
Human:
Do the sentiment analysis. Output "neu" for neutral sentence, "pos" for positive sentence ,  and "neg" for negative sentence. No explanation is required.
Sentence: {}
Assistant:

\subsubsection{Llama 3.1}
Perform sentiment analysis. Your output should be "pos" for positive sentence , "neu" for neutral sentence and "neg" for negative sentence. No explanation is required.

\subsubsection{Ministral 8B}
You are a helpful assistant in sentiment analysis. You should always provide answer from given labels without explanation.
Tweet: {}.  Perform sentiment analaysis on the tweet and answer with one of the following label:
-2 for  Highly negative
-1 for Negative
0 for Neutral
1 for Positive
2 for Highly positive
No explanation is required. 
Output (-2,-1,0,1,2):  

\subsection{Prompts - Sentiment Analysis (CLE)}
\subsubsection{Bloomz}
Your task is to perform sentiment analysis on the tweets. Labels are:
-2 : Highly negative
-1 : Negative
0 : Neutral
1 : Positive
2 : Highly positive
Output only label name. no explanation is required. Tweet {} . Output label:

\subsubsection{GPT 3.5 }
"system": "You are an expert in sentiment analysis on urdu tweets
Your task is to perform sentiment analysis on the tweets. Labels are:
-2 : Highly negative
-1 : Negative
0 : Neutral
1 : Positive
2 : Highly positive
Output only label name. no explanation is required. Tweet {} . Output label:

\subsubsection{Llama 2}

You are a helpful assistant in sentiment analysis. You should always provide answer from given labels without explanation.

Human:
Tweet: {}.  Perform sentiment analysis on the tweet and answer with one of the following label:
-2 : Highly negative
-1 : Negative
0 : Neutral
1 : Positive
2 : Highly positive
Assitant:

\subsubsection{Llama 3.1}
You are a helpful assistant in performing sentiment analysis. Perform sentiment analaysis on the tweet and answer with one of the following label:
-2 for  Highly negative
-1 for  Negative
0 for Neutral
1 for Positive
2 for Highly positive
Only output -2 to 2 based on the above mentioned scale. No explanation is required

\subsubsection{Ministral 8B}
You are a helpful assistant in performing sentiment analysis. Perform sentiment analaysis on the tweet and answer with one of the following label:
-2 for  Highly negative
-1 for  Negative
0 for Neutral
1 for Positive
2 for Highly positive
Only output -2 to 2 based on the above mentioned scale. No explanation is required

\subsection{Prompts - Multi-label Emotion Classification}
\subsubsection{Bloomz}
Output the emotion or emotions(if multiple) for the sentence. {}
Emotions: anger, disgust, fear, sadness, surprise, happiness, neutral. You can output multiple emotions as well but should only be the name of the emotions. Output:

\subsubsection{GPT 3.5 }
Output the emotion or emotions(if multiple) for the sentence. {}
Emotions: anger, disgust, fear, sadness, surprise, happiness, neutral. You can output multiple emotions as well but should only be the name of the emotions. Output:

\subsubsection{Llama 2}
Output the emotion or emotions(if multiple) for the sentence. {}
Emotions: anger, disgust, fear, sadness, surprise, happiness, neutral. You can output multiple emotions as well but should only be the name of the emotions. Output:

\subsubsection{Llama 3.1}
Output the emotion or emotions(if multiple) for the sentence.
Emotions: anger, disgust, fear, sadness, surprise, happiness, neutral. You can output multiple emotions as well but should only be the name of the emotions. No explanation is required.

\subsubsection{Ministral 8B}
Output the emotion or emotions for the paragraph. {}
Emotions: neutral, happy, fear, sad, anger, love. Your output should only be the name of one of the emotions. Output:

\subsection{Prompts - Emotion Classification }
\subsubsection{Bloomz}
Output the emotion or emotions for the paragraph. {}
Emotions: neutral, happy, fear, sad, anger, love. Your output should only be the name of one of the emotions. Output:

\subsubsection{GPT 3.5 }
"system", "content": "You are an expert in emotion recognition in the urdu samples "
Output the emotion or emotions for the paragraph. {}
Emotions: neutral, happy, fear, sad, anger, love. Your output should only be the name of one of the emotions. Output:

\subsubsection{Llama 2}
Output the emotion or emotions for the paragraph. {}
Emotions: neutral, happy, fear, sad, anger, love. Your output should only be the name of only one of the emotions. Output:

\subsubsection{Llama 3.1}
Output the emotions for the paragraph from one of the following:
neutral, happy, fear, sad, anger, love. 
Your output should only be the name of one of the given emotions. Don't provide any other apart from these six emotions. No explanation is required 

\subsubsection{Ministral 8B}
Output the emotions for the paragraph from one of the following:
neutral, happy, fear, sad, anger, love. 
Your output should only be the name of one of the given emotions. Don't provide any other apart from these six emotions. No explanation is required 

\subsection{Prompts - Machine Translation }

\label{subsec:MT}

\subsubsection{Bloomz}
You are an expert translator specialized in translating texts from English to Urdu .Translate the following English sentence to Urdu: {}

\subsubsection{GPT 3.5}
"system": "You are an expert translator specialized in translating texts from English to Urdu "
Translate the following English sentence to Urdu:{}

\subsubsection{Llama 2}
No explanation or notes required. Just translate. English: {} Urdu:

\subsubsection{Llama 3.1}
You are an English to Urdu translator. Translate the english sentences into Urdu. No explanation is required. Just translate into Urdu

\subsubsection{Ministral 8B}
You are an expert translator specialized in translating texts from English to Urdu. Translate the following English sentence to Urdu: "{}". Provide only the Urdu translation, without any additional text or explanations.

\subsection{Prompts - POS Tagging }

\label{subsec:MT}

\subsubsection{Bloomz}
Your task is to tag POS in input. You will use following Taggig scheme: Tag Proper Noun as NNP ,Tag Common Noun as NN,Tag Personal pronoun as PRP,Tag Demonstrative as PDM,Tag Possessive pronouns as PRS,Tag Reflexive pronouns as PRF,Tag Reflexive Apna as APNA,Tag Relative Personal as PRR,Tag Relative Demonstrative as PRD,Tag Main Verb Infinitive as VBI,Tag Main Verb Finite as VB,Tag Aspectual auxiliaries as AUXA,Tag Progressive auxiliaries as AUXP,Tag Tense auxiliaries as AUXT,Tag Modals auxiliaries as AUXM,Tag Foreign Fragment as FF,Tag Interjection as INJ,Tag Preposition as PRE,Tag Postposition as PSP,Tag Common as SYM,Tag Punctuation as PU,Tag Common as RB,Tag Negation as NEG,Tag Common as PRT,Tag Vala as VALA,Tag Coordinate Conjunction as CC,Tag Subordinate Conjunction as SC,Tag SC Kar as SCK,Tag Pre-sentential as SCP,Tag Ordinal as OD,Tag Fraction as FR,Tag Multiplicative as QM,Tag Adjective as JJ,Tag Quantifier as Q,Tag Cardinal as CD. Your Output should be only one tag corrosponding to input word. no explaination is required.  input: {}

\subsubsection{GPT 3.5}
"system": "You are an expert in Urdu pos tagging "
Your task is to tag POS in input. You will use following Taggig scheme: Tag Proper Noun as NNP ,Tag Common Noun as NN,Tag Personal pronoun as PRP,Tag Demonstrative as PDM,Tag Possessive pronouns as PRS,Tag Reflexive pronouns as PRF,Tag Reflexive Apna as APNA,Tag Relative Personal as PRR,Tag Relative Demonstrative as PRD,Tag Main Verb Infinitive as VBI,Tag Main Verb Finite as VB,Tag Aspectual auxiliaries as AUXA,Tag Progressive auxiliaries as AUXP,Tag Tense auxiliaries as AUXT,Tag Modals auxiliaries as AUXM,Tag Foreign Fragment as FF,Tag Interjection as INJ,Tag Preposition as PRE,Tag Postposition as PSP,Tag Common as SYM,Tag Punctuation as PU,Tag Common as RB,Tag Negation as NEG,Tag Common as PRT,Tag Vala as VALA,Tag Coordinate Conjunction as CC,Tag Subordinate Conjunction as SC,Tag SC Kar as SCK,Tag Pre-sentential as SCP,Tag Ordinal as OD,Tag Fraction as FR,Tag Multiplicative as QM,Tag Adjective as JJ,Tag Quantifier as Q,Tag Cardinal as CD. Your Output should be only one tag corrosponding to input word. no explaination is required.  input: {}

\subsubsection{Llama 2}
Your task is to tag POS in input. You will use following Taggig scheme: Tag Proper Noun as NNP ,Tag Common Noun as NN,Tag Personal pronoun as PRP,Tag Demonstrative as PDM,Tag Possessive pronouns as PRS,Tag Reflexive pronouns as PRF,Tag Reflexive Apna as APNA,Tag Relative Personal as PRR,Tag Relative Demonstrative as PRD,Tag Main Verb Infinitive as VBI,Tag Main Verb Finite as VB,Tag Aspectual auxiliaries as AUXA,Tag Progressive auxiliaries as AUXP,Tag Tense auxiliaries as AUXT,Tag Modals auxiliaries as AUXM,Tag Foreign Fragment as FF,Tag Interjection as INJ,Tag Preposition as PRE,Tag Postposition as PSP,Tag Common as SYM,Tag Punctuation as PU,Tag Common as RB,Tag Negation as NEG,Tag Common as PRT,Tag Vala as VALA,Tag Coordinate Conjunction as CC,Tag Subordinate Conjunction as SC,Tag SC Kar as SCK,Tag Pre-sentential as SCP,Tag Ordinal as OD,Tag Fraction as FR,Tag Multiplicative as QM,Tag Adjective as JJ,Tag Quantifier as Q,Tag Cardinal as CD. Your Output should be only one tag corrosponding to input word. no explaination is required.  input: {}

\subsubsection{Llama 3.1}
Your task is to tag POS in input. You will use following Taggig scheme: Tag Proper Noun as NNP ,Tag Common Noun as NN,Tag Personal pronoun as PRP,Tag Demonstrative as PDM,Tag Possessive pronouns as PRS,Tag Reflexive pronouns as PRF,Tag Reflexive Apna as APNA,Tag Relative Personal as PRR,Tag Relative Demonstrative as PRD,Tag Main Verb Infinitive as VBI,Tag Main Verb Finite as VB,Tag Aspectual auxiliaries as AUXA,Tag Progressive auxiliaries as AUXP,Tag Tense auxiliaries as AUXT,Tag Modals auxiliaries as AUXM,Tag Foreign Fragment as FF,Tag Interjection as INJ,Tag Preposition as PRE,Tag Postposition as PSP,Tag Common as SYM,Tag Punctuation as PU,Tag Common as RB,Tag Negation as NEG,Tag Common as PRT,Tag Vala as VALA,Tag Coordinate Conjunction as CC,Tag Subordinate Conjunction as SC,Tag SC Kar as SCK,Tag Pre-sentential as SCP,Tag Ordinal as OD,Tag Fraction as FR,Tag Multiplicative as QM,Tag Adjective as JJ,Tag Quantifier as Q,Tag Cardinal as CD. Your Output should be only one tag corrosponding to input word. no explaination is required.  input: {}

\subsubsection{Ministral 8B}
Your task is to tag POS in input. You will use following Taggig scheme: Tag Proper Noun as NNP ,Tag Common Noun as NN,Tag Personal pronoun as PRP,Tag Demonstrative as PDM,Tag Possessive pronouns as PRS,Tag Reflexive pronouns as PRF,Tag Reflexive Apna as APNA,Tag Relative Personal as PRR,Tag Relative Demonstrative as PRD,Tag Main Verb Infinitive as VBI,Tag Main Verb Finite as VB,Tag Aspectual auxiliaries as AUXA,Tag Progressive auxiliaries as AUXP,Tag Tense auxiliaries as AUXT,Tag Modals auxiliaries as AUXM,Tag Foreign Fragment as FF,Tag Interjection as INJ,Tag Preposition as PRE,Tag Postposition as PSP,Tag Common as SYM,Tag Punctuation as PU,Tag Common as RB,Tag Negation as NEG,Tag Common as PRT,Tag Vala as VALA,Tag Coordinate Conjunction as CC,Tag Subordinate Conjunction as SC,Tag SC Kar as SCK,Tag Pre-sentential as SCP,Tag Ordinal as OD,Tag Fraction as FR,Tag Multiplicative as QM,Tag Adjective as JJ,Tag Quantifier as Q,Tag Cardinal as CD. Your Output should be only one tag corrosponding to input word. no explaination is required.  input: {}

\end{document}